\documentclass{article}                     
\usepackage{cite}
\usepackage{natbib}    
\usepackage{color}
\usepackage{graphicx}
\usepackage{amsmath}
\usepackage{amssymb}
\usepackage{verbatim}
\usepackage{algorithm}
\usepackage[noend]{algorithmic}
\usepackage{subfigure}
\usepackage{url}
\usepackage{verbatim} 

\usepackage[colorlinks]{hyperref}
\usepackage{times}

\def\xx{{\boldsymbol{x}}}
\def\ww{{\boldsymbol{w}}}
\def\ff{{\boldsymbol{f}}}

\def\mmu{{\boldsymbol{\mu}}}
\renewcommand{\Re}{{\mathbb{R}}}

\newcommand{\argmin}[1]{{\hbox{$\underset{#1}{\mbox{argmin}}\;$}}}
\newcommand{\argmax}[1]{{\hbox{$\underset{#1}{\mbox{argmax}}\;$}}}

\begin{document}

\title{Segmentation of Natural Images by Texture and Boundary Compression}

\author{Hossein~Mobahi \and
 Shankar~R.~Rao \and
  Allen~Y.~Yang    \and
 S.~Shankar~Sastry \and
 Yi~Ma
}

\maketitle

\begin{abstract}
We present a novel algorithm for segmentation of natural images that harnesses the principle of minimum description length (MDL). Our method is based on observations that a homogeneously textured region of a natural image can be well modeled by a Gaussian distribution and the region boundary can be effectively coded by an adaptive chain code. The optimal segmentation of an image is the one that gives the shortest coding length for encoding all textures and boundaries in the image, and is obtained via an agglomerative clustering process applied to a hierarchy of decreasing window sizes as multi-scale texture features. The optimal segmentation also provides an accurate estimate of the overall coding length and hence the true entropy of the image. We test our algorithm on the publicly available Berkeley Segmentation Dataset. It achieves state-of-the-art segmentation results compared to other existing methods. 
\end{abstract}

{\small
{\hrule width 4.9cm}
\vspace{0.2cm}
\noindent Research was supported in part by NSF IIS 07-03756, ONR N00014-09-1-0230, ARO MURI W911NF-06-1-0076, and ARL MAST-CTA W911NF-08-2-0004. The views and conclusions contained in this document are those of the authors and should not be interpreted as representing the official policies, either expressed or implied, of the Army Research Laboratory or the U.S. Government. The U.S. Government is authorized to reproduce and distribute for Government purposes notwithstanding any copyright notation hereon.
\\
{\hrule width 4.9cm}
\vspace{0.2cm}
\noindent H.~Mobahi and Y.~Ma\\
Coordinated Science Lab, University of Illinois, Urbana, IL 61801, USA.\\
E-mail: \href{mailto:hmobahi2@illinois.edu}{hmobahi2@illinois.edu} , \href{mailto:yima@illinois.edu}{yima@illinois.edu}\\
\vspace{-0.2cm}
\\
S.~Rao\\
HRL Laboratories, LLC, Malibu, CA 90265, USA.\\
E-mail: \href{mailto:srrao@hrl.com}{srrao@hrl.com}\\
\vspace{-0.2cm}
\\
A.~Yang and S.~Sastry\\
Cory Hall, Department of EECS, University of California, Berkeley, CA 94720, USA.\\
E-mail: \href{mailto:yang@eecs.berkeley.edu}{yang@eecs.berkeley.edu} ,  \href{mailto:sastry@eecs.berkeley.edu}{sastry@eecs.berkeley.edu}\\
\vspace{-0.2cm}
\\
Y.~Ma\\
Visual Computing Group, Microsoft Research Asia, Beijing, China.
}

\pagebreak

\section{Introduction} 
\label{sec:introduction}
The task of partitioning a natural image into regions with homogeneous texture, commonly referred to as {\em image segmentation}, is widely accepted as a crucial function for high-level image understanding, significantly reducing the complexity of content analysis of images.
Image segmentation and its higher-level applications are largely designed to emulate functionalities of human visual perception (e.g., in object recognition and scene understanding). 
Dominant criteria for measuring segmentation performance are based on qualitative and quantitative comparisons with human segmentation results. In the literature, investigators have explored several important models and principles that can lead to good image segmentation:
 
\begin{enumerate}
\item Different texture regions of a natural image admit a mixture model. For example, Multiscale Normalized Cuts (MNC) by \cite{CourT2005-CVPR} and F\&H by \cite{FelzenszwalbP2004-IJCV} formulate the segmentation as a graph-cut problem, while Mean Shift (MS) by \cite{ComaniciuD2002-PAMI} seeks a partition of a color image based on different modes within the estimated empirical distribution.
\item Region contours/edges convey important information about the saliency of the objects in the image and their shapes (see \cite{ElderJ1996-ECCV,ArbelaezP2006,ZhuQ2007-ICCV,RenX2008-IJCV}). Several recent methods have been proposed to combine the cues of homogeneous color and texture with the cue of contours in the segmentation process, including \cite{MalikJ2001-IJCV,TuZ2002-PAMI,KimJ2005-PAMI}.
\item The properties of local features (including texture and edges) usually do not share the same level of homogeneity at the same spatial scale. Thus, salient image regions can only be extracted from a hierarchy of image features under multiple scales (see \cite{YuS2005-CVPR,RenX2005-ICCV,YangA2008-CVIU}). 
\end{enumerate} 

Despite much work in this area, good image segmentation remains elusive to obtain for practitioners, mainly for the following two reasons: First, there is little consensus on what criteria should be used to evaluate the quality of image segmentations. It is difficult to strike a good balance between objective measures that depend solely on the intrinsic statistics of imagery data and subjective measures that try to empirically mimic human perception. Second, in the search for objective measures, there has been a lack of consensus on good models for a unified representation of image segments including both their textures and contours. 

Recently, an {\em objective} metric based on the notion of lossy {\em minimum description length} (MDL) has been proposed for evaluating clustering of general mixed data (\cite{MaY2007-PAMI}). The basic idea is that, given a potentially mixed data set, the ``optimal segmentation'' is the one that, over all possible segmentations, minimizes the coding length of the data, subject to a given quantization error. For data drawn from a mixture of Gaussians, the optimal segmentation can often be found efficiently using an agglomerative clustering approach. The MDL principle and the new clustering method have later been applied to the segmentation of natural images, known as  {\em compression-based texture merging} (CTM) (\cite{YangA2008-CVIU}). This approach has proven to be highly effective for imitating human segmentation of natural images. Preliminary success of this approach leads to the following important question: \emph{To what extent is segmentation obtained by image compression consistent with human perception?}

The CTM algorithm also has its drawbacks. In particular, although the CTM method utilizes the idea of data compression, it does not exactly seek to compress the image {\em per se}. First, it ``compresses'' feature vectors or windows extracted around all pixels by grouping them into clusters as a mixture of Gaussian models. As a result, the final coding length is highly {\em redundant} due to overlapping between windows of adjacent pixels, and has no direct relation to the true entropy of the image. Second, the segmentation result encodes the membership of pixels using a Huffman code, which does not take into account the smoothness of boundaries nor the spatial relationship of adjacent pixels that are more likely belong to one texture region. Thus, CTM does not give a good estimate of the true entropy of the image, and it cannot be used to justify a strong connection between image segmentation and image compression. 
   
\subsection{Contributions}
In this paper, we contend that, much better segmentation results can be obtained if we more closely adhere to the principle of image compression, by correctly counting only the necessary bits needed to encode a natural image for both the texture and boundaries. The proposed algorithm precisely estimates the coding length needed to encode the texture of each region based on the rate distortion of its probabilistic distribution and the number of \emph{non-overlapping} interior windows. In order to adapt to different scales and shapes of texture regions in an image, a hierarchy of multiple window sizes is incorporated in the segmentation process. The algorithm further encodes the boundary information of each homogeneous texture region by carefully counting the number of bits needed to encode the boundary with an adaptive chain code.

Based on the MDL principle, the optimal segmentation of an image is defined as the one that minimizes its total coding length, in this case a close approximation to the true entropy of the image. With any fixed quantization, the final coding length gives a purely objective measure for how good the segmentation is in terms of the level of image compression. Finally, we propose a simple yet effective regression method to adaptively select a proper quantization level for each individual image to achieve the optimal segmentation result.

We conduct extensive experiments to compare the results with human segmentation using the Berkeley Segmentation Dataset (BSD) (\cite{MartinD2001-ICCV}). Although our method is conceptually simple and the measure used is purely objective, the segmentation results match extremely well with those by humans, exceeding or competing with the best segmentation algorithms.  

\section{Coding Length Functions for Texture and Boundary}
\label{sec:compression}
\subsection{Construction of Texture Features}
\label{sec:Gaussianity}
We first discuss how to construct texture vectors that represent homogeneous textures in image segments. In order to capture the variation of a local \emph{texton}, one can directly apply a $w\times w$ cut-off window around a pixel across the three color channels, and stack the color values inside the window in a vector form as in \cite{YangA2008-CVIU}. \footnote{Another popular approach for constructing texture vectors is to use multivariate responses of a fixed 2-D texture filter bank. A previous study by \cite{VarmaM2003} has argued that the difference in segmentation results between the two approaches is small, and yet it is more expensive to compute 2-D filter bank responses.}  
\begin{figure}[hbt]
\centering
\includegraphics[width=2.5in]{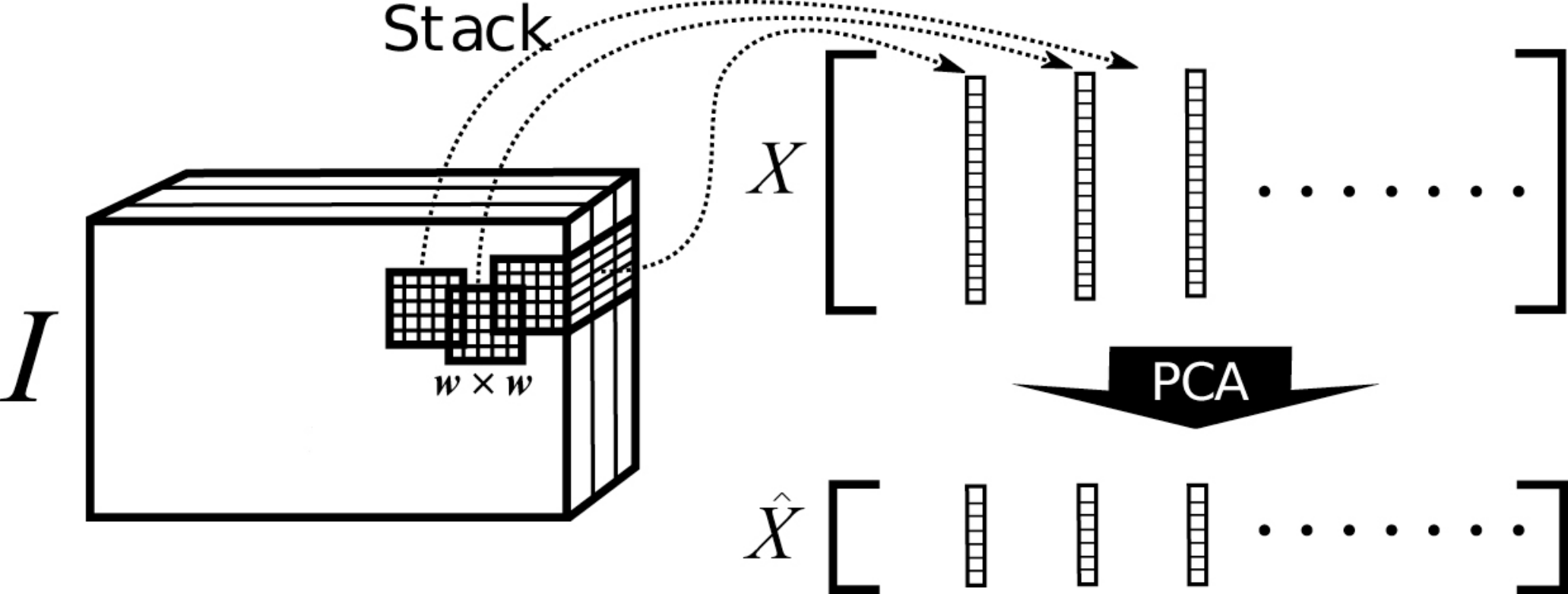} 
\caption{\small Texture features are constructed by stacking the $w \times w$ windows around all pixels of a color image $I$ into a data matrix $X$ and then projected to a low-dimensional space via principal component analysis (PCA). }
\label{fig:windows}   
\end{figure}

Figure \ref{fig:windows} illustrates the process of constructing texture features. Let the $w$-neighborhood $\mathcal{W}_w(p)$ be the set of all pixels in a $w \times w$ window across three color channels (e.g., $RGB$ or $L^*a^*b^*$) centered at pixel $p$. Define the set of features $X$ by taking the $w$-neighborhood around each pixel in $I$, and then stacking the window as a column vector:  
\begin{equation}
 X \doteq \{\xx_p\in \Re^{3w^2}: \; \xx_p = {\mathcal{W}_w(p)}^S \mbox{ for  } p \in I\}.   
 \end{equation}
For ease of computation, we further reduce the dimensionality of these features by projecting the set of all features $X$ onto their first $D$ principal components. We denote the set of features with reduced dimensionality as  $\hat{X}$. We have observed that for many natural images, the first eight principal components of $X$ contain over $99\%$ of the energy. In this paper, we choose to assign $D=8$.

Over the years, there have been many proposed methods to model the representation of image textures in natural images. One model that has been shown to be successful in encoding textures both
empirically and theoretically is the Gaussian {\em Mesh Markov Model} (MMM) \citep{LevinaE2006-AS}. Particularly in texture synthesis, the Gaussian MMM can provide consistent estimates of the joint distribution of the pixels in a window, which then can be used to fill in missing texture patches via a simple nonparametric scheme \citep{EfrosA1999-ICCV}.

However, to determine the optimal compression rate for samples from a distribution, one must know the rate-distortion function of that distribution \citep{YangA2008-CVIU}. Unfortunately, to our knowledge, the rate-distortion function for MMMs is not known in closed form and difficult to estimate empirically. Over all distributions with the same variance, it is known that the Gaussian distribution has the highest rate-distortion, and is in this sense the worst case distribution for compression. Thus by using the rate-distortion for a Gaussian distribution, we obtain an \emph{upper bound} for the true coding length of the MMM.

\subsection{Texture Encoding}
\label{sec:texture}
To describe encoding texture vectors, we first consider a single region $R$ with $N$ pixels. Based on \cite{YangA2008-CVIU}, for a fixed quantization error $\varepsilon$, the expected number of bits needed to code the set of $N$ feature windows $\hat{X}$ up to distortion $\varepsilon^2$ is given by: 
\begin{eqnarray}
L_{\varepsilon}(\hat{X}) &\doteq&   \underbrace{\tfrac{D}{2}\log_2 \det(I + \tfrac{D}{\varepsilon^2} \Sigma)}_{\mbox{codebook}} +  \underbrace{\tfrac{N}{2}\log_2 \det(I + \tfrac{D}{\varepsilon^2} \Sigma)}_{\mbox{data}}   + \underbrace{\tfrac{D}{2}\log_2(1 + \tfrac{\|\mmu\|^2}{\varepsilon^2})}_{\mbox{mean}}, \label{eqn:old-coding}  
\end{eqnarray}
where $\mmu$ and $\Sigma$ are the mean and covariance of the vectors in $\hat{X}$. Equation \eqref{eqn:old-coding} is the sum of three coding-lengths for the $D$ Gaussian principal vectors as the codebook, the $N$ vectors w.r.t. that codebook, and the mean of the Gaussian distribution.

\begin{figure}[ht!]
\centering
\subfigure[]{\includegraphics[width=2.5in]{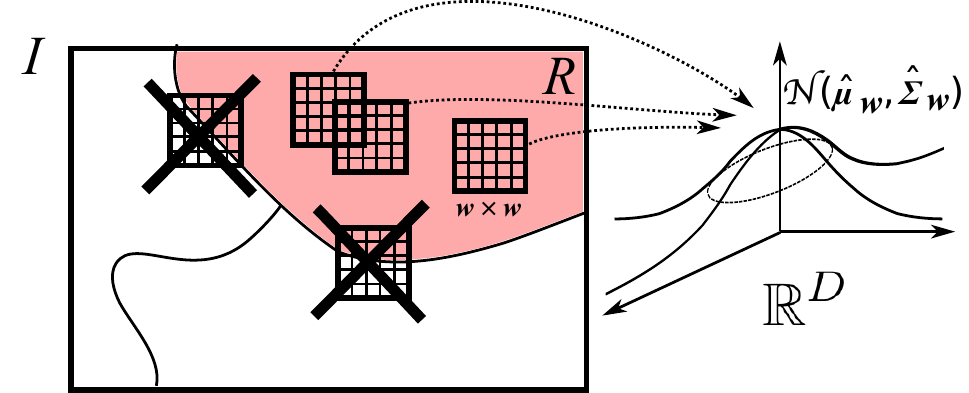}}\hspace{.1\textwidth}
\subfigure[]{\includegraphics[width=1.70in]{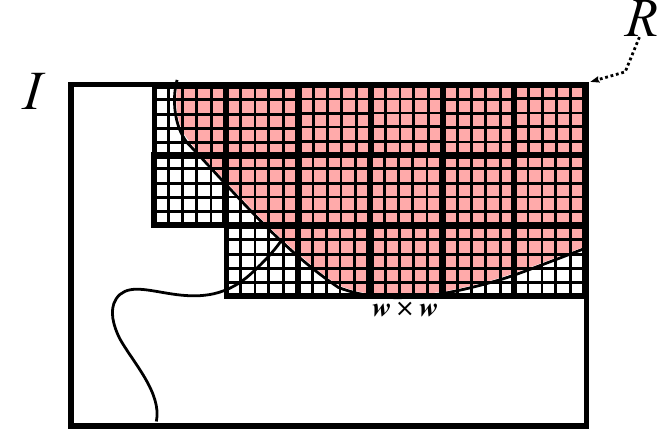}}
\caption{\small (a) Only windows from the interior of a region are used to compute the empirical mean $\hat{\mmu}_w$ and covariance $\hat{\Sigma}_w$. (b) Only nonoverlapping windows that can tile R as a grid are encoded. }
\label{fig:interior}   
\end{figure}

The coding length function \eqref{eqn:old-coding} is uniquely determined by the mean and covariance $(\mmu, \Sigma)$. To estimate them empirically, we need to exclude the windows that cross the boundary of $R$ (as shown in Figure \ref{fig:interior}(a)). Such windows contain textures from the adjacent regions, which cannot be well modeled by a single Gaussian as the interior windows. Hence, the empirical mean $\hat{\mmu}_w$ and covariance $\hat{\Sigma}_w$ of $R$ are only estimated from the {\em interior}  of $R$: 
\begin{equation}
\mathcal{I}_w(R) \doteq \{ p \in R : \forall q\in\mathcal{W}_w(p), q \in R\}.  
\end{equation} 

Furthermore, in \eqref{eqn:old-coding}, encoding all texture vectors in $\hat{X}$ to represent region $R$ is highly redundant because the $N$ windows {\em overlap} with each other. Thus, to obtain an efficient code of $R$ that closely approximates its true entropy, we only need to code the \emph{nonoverlapping} windows that can tile $R$ as a grid, as in Figure \ref{fig:interior} (b). 

Ideally, if $R$ is a rectangular region of size $mw \times nw$, where $m$ and $n$ are positive integers, then clearly we can tile $R$ with exactly $mn = \tfrac{N}{w^2}$ windows. So for coding the region $R$, \eqref{eqn:old-coding} becomes: 
 
\begin{eqnarray}\label{eqn:non-overlapping}
L_{w,\varepsilon}(R) &\doteq&   (\tfrac{D}{2} + \tfrac{N}{2w^2})\log_2 \det(I + \tfrac{D}{\varepsilon^2} \hat{\Sigma}_w)   +  \tfrac{D}{2}\log_2(1+ \tfrac{\|\hat{\mmu}_w\|^2}{\varepsilon^2}).  
\end{eqnarray}
Real regions in natural images  normally do not have such nice rectangular shapes. However, \eqref{eqn:non-overlapping} remains a good approximation to the actual coding length of a region $R$ with relatively smooth boundaries.\footnote{For a large region with a sufficiently smooth boundary, the number of boundary-crossing windows is significantly smaller than the number of those in the interior. For boundary-crossing windows, their average coding length is roughly proportional to the number of pixels inside the region if the Gaussian distribution is sufficiently isotropic.}

\subsection{Boundary Encoding}
\label{sec: boundary}

 To code windows from multiple regions in an image, one must know to which region each window belongs, so that each window can be decoded w.r.t. the correct codebook. For generic samples from multiple classes, one can estimate the distribution of each class label and then code the membership of the samples using a scheme that is asymptotically optimal for that class distribution (such as the Huffman code used in \cite{YangA2008-CVIU}). Such coding schemes are highly inefficient for natural image segmentation, as they do not leverage the spatial correlation of pixels in the same region. In fact, for our application, pixels from the same region form a connected component. Thus, the most efficient way of coding group membership for regions in images is to code the {\em boundary} of the region containing the pixels.

\begin{figure}[t]
\centering
\begin{minipage}[b]{1in}
\vspace{.1in}
\begin{tabular}[b]{|@{\,}c@{\,}c@{\,}c@{\,}c@{\,}c@{\,}|}
\hline
3 & & 2 & & 1\\
& $\nwarrow$ & $\uparrow$ & $\nearrow$ & \\
4 & $\leftarrow$ &  $\bullet$ & $\rightarrow$ & 0\\
& $\swarrow$ & $\downarrow$ & $\searrow$ & \\
5 & & 6 & &  7\\
\hline
\end{tabular}
\vspace{.2in}
\end{minipage}\hspace{.4in}
\includegraphics[width=1in]{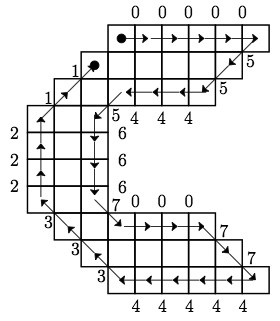}
\hspace{.2in}
\includegraphics[width=1in]{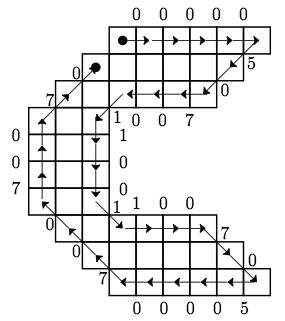}
 
\caption{\small {\bf Left: } The Freeman chain code of an edge orientation along 8 possible directions. {\bf Middle: } Representation of the boundary of a region in an image w.r.t. the Freeman chain code. {\bf Right:} Representation w.r.t the difference chain code. }
\label{fig:chain-codes}   
\end{figure}

A well-known scheme for representing boundaries of image regions is the {\em Freeman chain code}. In this coding scheme, the orientation of an edge is quantized along 8 discrete directions, shown in Figure \ref{fig:chain-codes}. Let $\{o_t\}_{t=1}^T$ denote the orientations of the $T$ boundary edges of $R$. Since each chain code can be encoded using three bits, the coding length of the boundary of $R$ is  
 
\begin{equation}
B(R) = 3\sum_{i=0}^7\#(o_t=i).
\label{eq:boundary-coding-length} 
\end{equation} 

The coding length $B(R)$ can be further improved by using an adaptive Huffman code that leverages the prior distribution of the chain codes. Though the distribution of chain codes is essentially uniform in most images, for regions with smooth boundaries, we expect that the orientations of consecutive edges are similar, and so consecutive chain codes will not differ by much. Given an initial orientation (expressed in chain code) $o_t$, the {\em difference chain code} of the following orientation $o_{t+1}$ is $\Delta o_t\doteq\mod(o_t-o_{t+1}, 8)$.  Figure \ref{fig:chain-codes} compares the original Freeman chain code with the difference chain code for representing the boundary of a region. Notice for this region, the difference encoding uses only half of the possible codes, with most being zeroes, while the Freeman encoding uses all eight chain codes. Given the prior distribution $P[\Delta o]$ of difference chain codes, $B(R)$ can be encoded more efficiently using a lossless Huffman coding scheme:
 
\begin{equation}
B(R) = -\sum_{i=0}^{7}\#(\Delta o_t=i)\log_2(P[\Delta o = i]).
\label{eq:boundary-predicted-coding-length} 
\end{equation}
For natural images, we estimate  $P[\Delta o]$ using images from the BSD  that were manually segmented by humans. We compare our distribution with one estimated by \cite{LiuY2005-PR}, who used 1000 images of curves, contour patterns, and shapes obtained from the web. As the results in Table \ref{tab:chain-code-prior} show, the regions of natural images tend to have more smooth boundaries when segmented by humans. 
\begin{table}[htb] 
\caption{\small The prior probability of the difference chain codes estimated from the BSD and by  \cite{LiuY2005-PR}.} 
\centering
\begin{small}
\begin{tabular}{|c||c|c|c|c|c|c|c|c|c|}
\hline
Difference Code & 0 & 1 & 2 & 3 & 4 & 5 & 6 & 7 \\
\hline
Angle change & $0^{\circ}$ & $45^{\circ}$ & $90^{\circ}$ & $135^{\circ}$ & $180^{\circ}$ & $-135^{\circ}$ & 
$-90^{\circ}$ & $-45^{\circ}$  \\
\hline
Probability (BSD) & 0.585 & 0.190 & 0.020 & 0.000 & 0.002 & 0.003 & 0.031 & 0.169 \\
\hline
Probability (Liu-Zalik) & 0.453 & 0.244 & 0.022 & 0.006 & 0.003 & 0.006 & 0.022 & 0.244 \\
\hline
\end{tabular}
\end{small}
\label{tab:chain-code-prior}   
\end{table}

\section{Image Segmentation Algorithm}
\label{sec:algorithm}

In this section, we discuss how to use the coding length functions to construct a better compression-based image segmentation algorithm. We first describe a basic approach. Then we propose a hierarchical scheme to deal with small regions using multi-scale texture windows. Finally, we investigate a simple yet effective regression scheme to adaptively choose a proper distortion parameter $\varepsilon$ based on a set of manually labeled segmentation examples.

\subsection{Minimization of the Total Coding Length Function}
Suppose an image $I$ can be segmented into non-overlapping regions \\
$\mathcal{R} = \{R_1, \ldots, R_k\},$ $\ \cup_{i=1}^k R_i=I$. The total coding length of the image $I$ is 
 
\begin{eqnarray}
L^S_{w,\varepsilon}(\mathcal{R}) &\doteq& \sum_{i=1}^k L_{w,\varepsilon}(R_i) + \tfrac{1}{2}B(R_i). \label{eqn:total-coding-length}   
\end{eqnarray}
Here, the boundary term is scaled by a half because we only need to represent the boundary between any two regions once. The optimal segmentation of $I$ is the one that minimizes \eqref{eqn:total-coding-length}.  Finding this optimal segmentation  is, in general, a combinatorial task, but we can often do so using an {\em agglomerative} approximation. 

To initialize the optimization process, one can assume each image pixel (and its windowed texture vector) belongs to an individual group of its own. However, this presents a problem that the maximal size of the texture window can only be one without intersecting with other adjacent regions (i.e., other neighboring pixels). In our implementation, similar to \cite{YangA2008-CVIU}, we utilize an oversegmentation step to initialize the optimization by \emph{superpixels}. A superpixel is a small region in the image that does not contain strong edges in its interior. Superpixels provide a coarser quantization of an image than the underlying pixels, while respecting strong edges between the adjacent homogeneous regions. There are several methods that can be used to obtain a superpixel initialization, including those of  \cite{MoriG2004-CVPR},  \cite{FelzenszwalbP2004-IJCV}, and \cite{RenX2005-ICCV}. We have found that \cite{MoriG2004-CVPR}\footnote{We use the publicly available code for this method available at \url{http://www.cs.sfu.ca/~mori/research/superpixels/} with parameter $\texttt{N\_sp} = 200$. } works well for our purposes. 

Given an oversegmentation of the image, at each iteration, we find the pair of regions $R_i$ and $R_j$ that will maximally decrease \eqref{eqn:total-coding-length} if merged:  
\begin{small}
 
\begin{eqnarray*}
\hspace{-4mm}(R_i^*, R_j^*) &=& \argmax{R_i, R_j \in \mathcal{R}} \Delta L_{w,\varepsilon}(R_i, R_j), \quad \mbox{where  \hspace{2mm}}  
\end{eqnarray*} 
\end{small}   
\begin{small}
\begin{eqnarray}
\hspace{-.05in}\Delta L_{w,\varepsilon}(R_i, R_j) &\doteq& L^S_{w,\varepsilon}(\mathcal{R})  - L^S_{w,\varepsilon}((\mathcal{R}\backslash \{R_i, R_j\}) \cup \{R_i\cup R_j\}) \nonumber \\
&=&  L_{w,\varepsilon}(R_i) + L_{w,\varepsilon}(R_j)   - L_{w,\varepsilon}(R_i \cup R_j)  \nonumber \\
&& + \tfrac{1}{2}(B(R_i) + B(R_j) - B(R_i \cup R_j)).  \label{eqn:delta-l}
 \end{eqnarray}
 \end{small}$\Delta L_{w,\varepsilon}(R_i, R_j)$ essentially captures the difference in the lossy coding lengths of the texture regions $R_i$ and $R_j$ and their boundaries before and after the merging. If $\Delta L(R_i^*, R_j^*) > 0$, we merge $R_i^*$ and $R_j^*$ into one region,  and repeat this process, continuing until the coding length $L^S_{w,\varepsilon}(\mathcal{R})$ can not be further reduced. 
 
To model the spatial locality of textures, we further construct a {\em region adjacency graph} (RAG): $\mathcal{G} = (\mathcal{V}, \mathcal{E})$. Each vertex $v_i \in\mathcal{V}$ corresponds to region $R_i \in \mathcal{R}$, and an edge $e_{ij} \in \mathcal{E}$ indicates that regions $R_i$ and $R_j$ are adjacent in the image. To perform image segmentation, we simply apply a constrained version of the above agglomerative procedure -- only merging regions that are adjacent in the image.

\subsection{Hierarchical Implementation}
\label{sec:hierarchy} 
\begin{figure}[htb]
\subfigure[Initial regions]{\includegraphics[width=.4\textwidth]{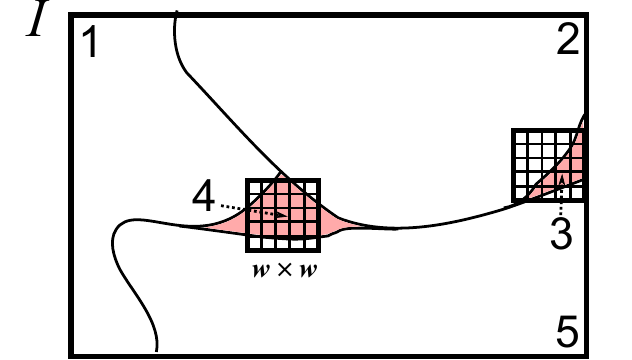}}
\subfigure[Stage 1]{\includegraphics[width=.4\textwidth]{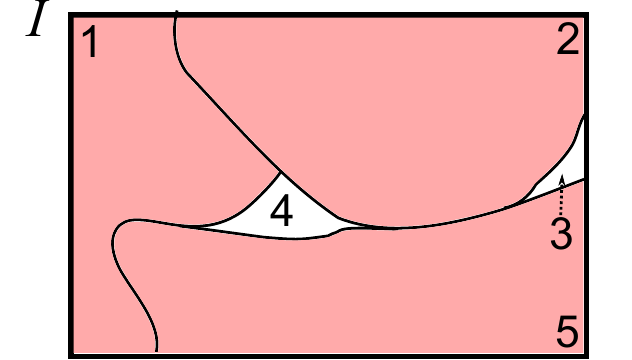}} 
\subfigure[Stage 2]{\includegraphics[width=.4\textwidth]{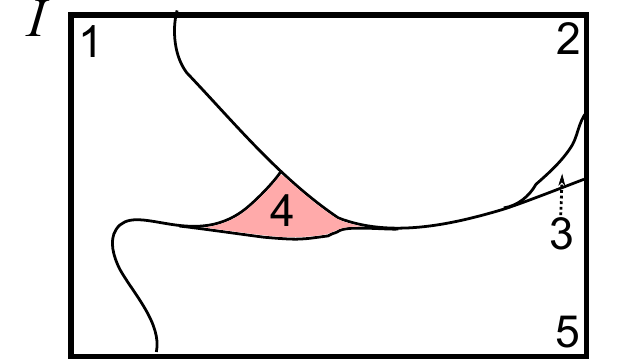}} \hspace{.185\textwidth}
\subfigure[Stage 3]{\includegraphics[width=.4\textwidth]{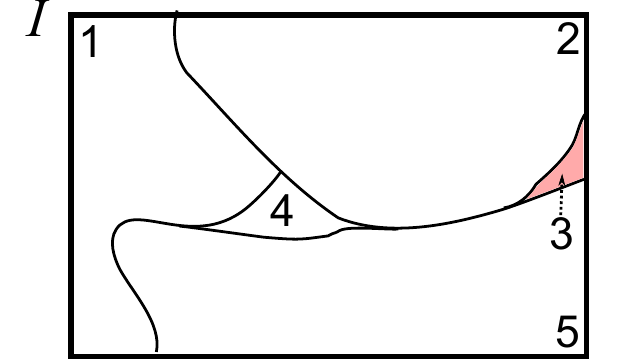}}
\caption{An example of our scheme for hierarchical image segmentation (a) Initial set of regions. Note that regions 3 and 4 are degenerate w.r.t. the window size $w$.  (b) In the first stage, only nondegenerate regions 1, 2, and 5 are considered for merging. (c) In the next stage, $w$ is reduced, causing region 4 to be marginally nondegenerate. We consider merging region 4 with it's nondegenerate neighbors. (d) ln a later stage, $w$ is reduced enough so that region 3 becomes nondegenerate. These stages are repeated until the overall coding length can no longer be decreased.} \label{fig:hierarchy}
\end{figure}

The above region-merging scheme is based on the assumption of a fixed texture window size, and clearly cannot effectively deal with regions or superpixels that are very small. In such cases, the majority of the texture windows will intersect with the boundary of the regions. We say that a region $R$ is {\em degenerate } w.r.t. window size $w$ if $\mathcal{I}_w(R) = \emptyset$. For such regions, the $w$-neighborhoods of all pixels will contain pixels from other regions, and so $\hat{\mmu}$ and $\hat{\Sigma}$ cannot be reliably estimated. These regions are degenerate precisely because of the window size; for any $w$-degenerate region $R$, there is $1 \le w' < w$ such that $\mathcal{I}_{w'}(R) \ne \emptyset$. We say that $R$ is {\em marginally nondegenerate} w.r.t. window size $w$ if $\mathcal{I}_w(R) \ne \emptyset$ and $\mathcal{I}_{w+2}(R) = \emptyset$. To deal with these degenerate regions, we propose to use a hierarchy of  window sizes. Starting from the largest window size, we recursively apply the above scheme with ever smaller window sizes till all degenerate regions have been merged with their adjacent ones. In this paper, we start from $7\times 7$ and reduce to $5 \times 5, 3\times 3$, and $1\times 1$. Please refer to Figure \ref{fig:hierarchy} for an example
of our hierarchical scheme. 

Notice that at a fixed window size, the region-merging process is similar to the CTM approach proposed in \cite{YangA2008-CVIU}. Nevertheless, the new coding length function and the hierarchical implementation give much more accurate approximation to the true image entropy and hence lead to much better segmentation results. We summarize the overall algorithm for image segmentation in Algorithm \ref{alg:multi-resolution}, which we refer to as {\em Texture and Boundary Encoding-based Segmentation} (TBES).
\begin{algorithm}[htb]
\caption{\bf Texture and Boundary Encoding-based Segmentation (TBES)}
\footnotesize Given image $I$, distortion $\varepsilon$, max window size $w_M$, superpixels $\mathcal{R} = \{R_1, \ldots, R_k \}$,
\begin{algorithmic}[1]
\FOR{$w = 1:2:w_M$}
\STATE Construct $\hat{X}_w$ by stacking the $w \times w$ windows around each $p \in I$ as column vectors and applying PCA.
\ENDFOR
\STATE Construct RAG $\mathcal{G} = (\mathcal{V}, \mathcal{E})$, where $\mathcal{V} \simeq \mathcal{R}$ and $e_{ij} \in \mathcal{E}$ only if $R_i$ and $R_j$ are adjacent in $I$.
\STATE $w = w_M$
\REPEAT
\IF{$w = w_M$}
\STATE Find $R_i$ and $R_j$ such that $e_{ij} \in \mathcal{E}$,  $\mathcal{I}_w(R_i) \ne \emptyset$, $\mathcal{I}_w(R_j) \ne \emptyset$, and
$\Delta L_{w,\varepsilon}(R_i,R_j) $
 is maximal.
 \ELSE \STATE Find  $R_i$ and   $R_j$ such that $e_{ij} \in \mathcal{E}$,  $\mathcal{I}_w(R_i) \ne \emptyset$, $\mathcal{I}_w(R_j)\ne \emptyset$ , $\mathcal{I}_{w+2}(R_i) = \emptyset$ or $\mathcal{I}_{w+2}(R_j) = \emptyset$ and $\Delta L_{w,\varepsilon}(R_i,R_j) $
 is maximal.
\ENDIF 
\IF{$\Delta L_{w,\varepsilon}(R_i,R_j) > 0$}
\STATE $\mathcal{R} := \left(\mathcal{R} \ \backslash\ \{R_i, R_j\}\right) \cup \{R_i \cup R_j\}.$
\STATE Update $\mathcal{G}$  based on the newly merged region.
\STATE $w = w_M$
\ELSIF{$w \ne 1$}
\STATE $w = w - 2$
\ENDIF
\UNTIL{$ \mathcal{I}_{w_M}(R)\ne\emptyset, \ \ \  \forall R \in \mathcal{R} \ \ \ $ {\bf and} $ \ \ \  \Delta L_{w_M,\varepsilon}(R_i,R_j) \le 0, \ \ \ \ \forall R_i,R_j \in \mathcal{R}$ }
 \STATE {\bf Output: } The set of regions  $\mathcal{R}$.
\end{algorithmic}
\label{alg:multi-resolution}
\end{algorithm}

\subsection{Choosing the Distortion Level}
\label{sec:epsilon}

\begin{figure}[ht!]
\centering
\subfigure[Original images.]{
\includegraphics[width=1in]{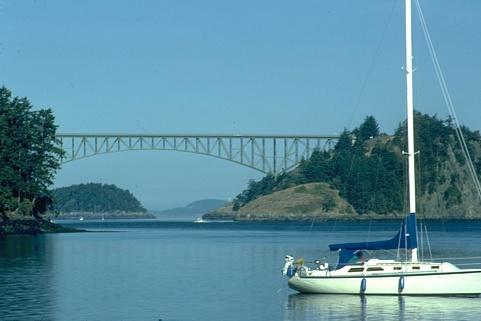} 
\includegraphics[width=1in]{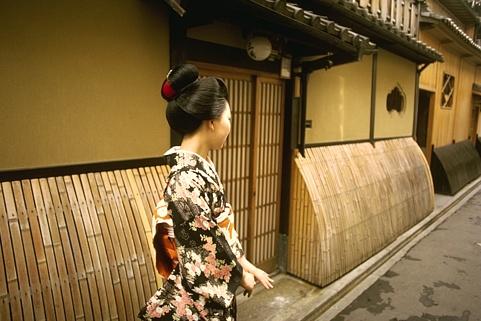} 
\includegraphics[width=1in]{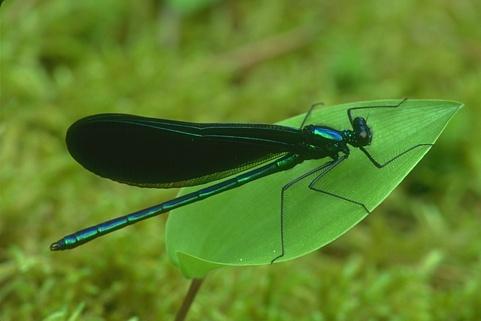} 
\includegraphics[width=1in]{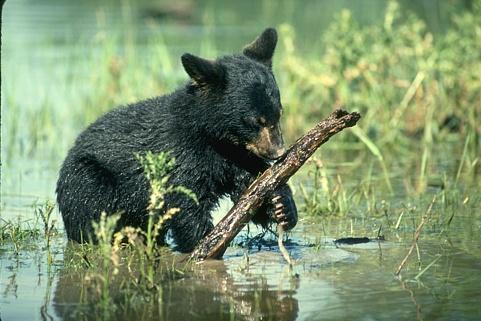} 
}
\subfigure[Segmentation results with distortion ($\varepsilon=25$)]{
\includegraphics[width=1in]{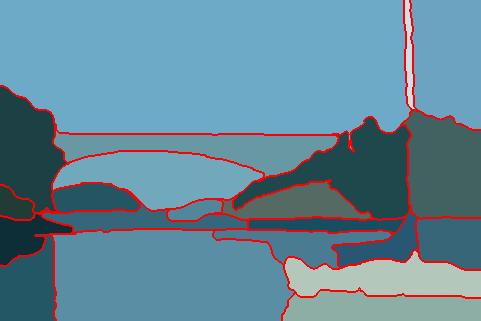}
\includegraphics[width=1in]{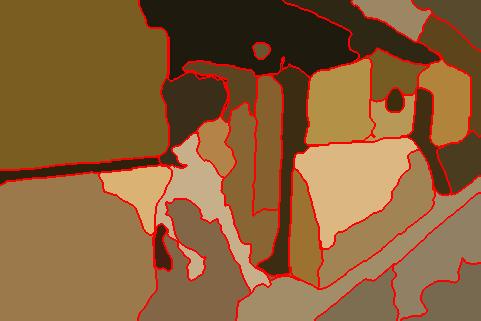}
\includegraphics[width=1in]{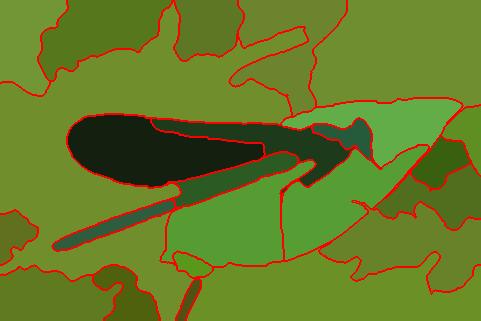}
\includegraphics[width=1in]{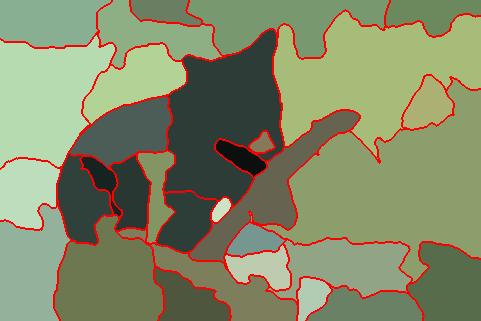}
}
\subfigure[Segmentation results with distortion ($\varepsilon=400$)]{
\includegraphics[width=1in]{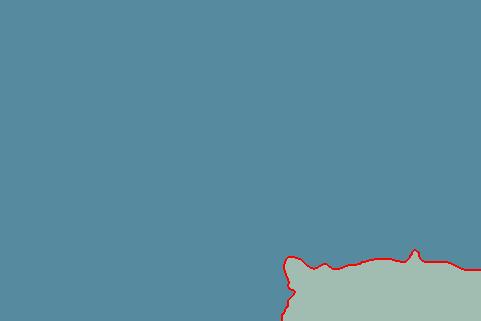} 
\includegraphics[width=1in]{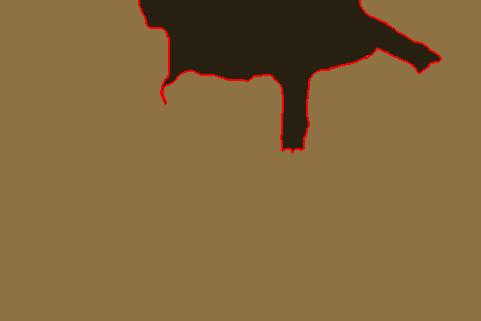} 
\includegraphics[width=1in]{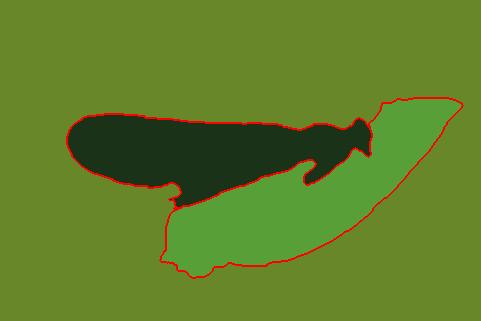} 
\includegraphics[width=1in]{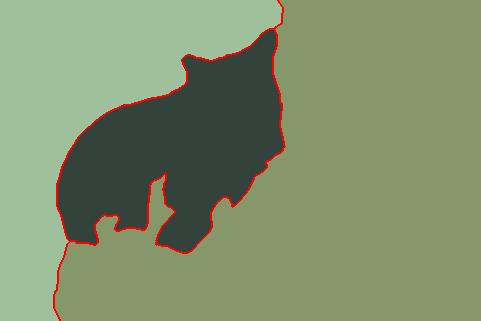} 
}
\caption{\small A comparison of segmentation results w.r.t. different distortion levels. The low distortion generates better segmentations for the left two images, while the high distortion 
generates better results for the right two images.}
\label{fig:different-eps}  
\end{figure}

Algorithm \ref{alg:multi-resolution} requires a single parameter, the distortion level $\varepsilon$, that determines the coarseness of the segmentation. The optimality of $\varepsilon$ is measured by the segmentation that best matches with human perception. As shown in Figure \ref{fig:different-eps}, since natural images have different scales of resolution, no single choice of $\varepsilon$ is optimal for all images. In this section, we propose a solution to adaptively select a proper distortion parameter such that the segmentation result better approximates human perception. The method assumes that a set of training images $\mathcal{I}=\{I_k\}$ have been manually segmented by human users as the ground truth set $\mathcal{S}_g=\{\mathcal{R}_g(I_k)\}$.

\begin{figure}[ht!]
\centering
\subfigure[Input Image]{
\includegraphics[height=2in]{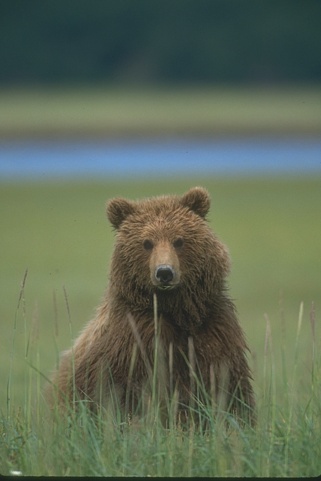} 
} \quad \quad
\subfigure[Distortion vs Discrepancy of Segmentation]{
\includegraphics[height=2.2in]{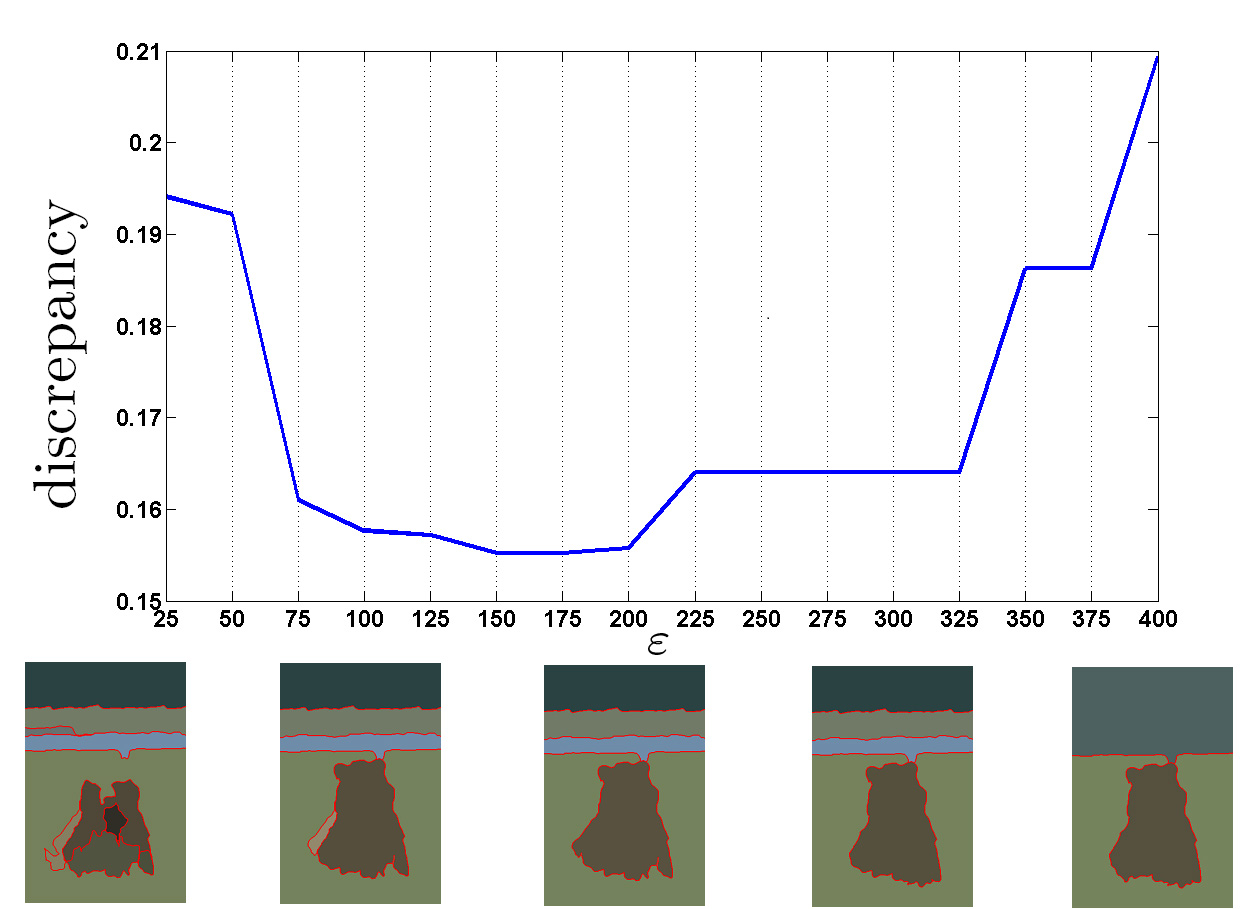} 
}
\caption{\small The effect of distortion $\varepsilon$ on the discrepancy $d(\mathcal{R}_\varepsilon(I_k),\mathcal{R}_g(I_k))$ on an example image. The discrepancy shown in the plot is the probability that an arbitrary pair of pixels do not have consistent labels in $\mathcal{R}_\varepsilon(I_k)$ and $\mathcal{R}_g(I_k)$, namely, $\mbox{PRI}^C$ (please refer to Section \ref{sec:setup}).}
\label{fig:epsilon-vs-discrepancy}  
\end{figure}

To objectively quantify how well a given segmentation matches with human perception, we first need a measure for the discrepancy between two segmentations $\mathcal{R}_1$ and $\mathcal{R}_2$, denoted as $d(\mathcal{R}_1, \mathcal{R}_2)$. Intuitively, the discrepancy measure should be small when $\mathcal{R}_1$ and $\mathcal{R}_2$ are similar in some specific sense.\footnote{We will discuss several discrepancy measures in Section \ref{sec:setup}, such as the probabilistic Rand index (PRI) and variation of information (VOI).} Given a measure $d$, the best $\varepsilon$ for $I_k$, denoted by $\varepsilon^*_k$, can be obtained by:
\begin{equation}
\varepsilon_k^* = \arg\min_{\varepsilon} d(\mathcal{R}_\varepsilon(I_k),\mathcal{R}_g(I_k)), \quad \mbox{for each } I_k\in \mathcal{I}.
\label{eq:optimal-epsilon-estimate}
\end{equation}
An example of the relationship between $\varepsilon$ and a discrepancy measure $d$ is shown in Figure \ref{fig:epsilon-vs-discrepancy}.

As ground truth segmentations are not available for non-training images, we shall use the training images $\mathcal{S}_g=\{\mathcal{R}_g(I_k)\}$ to infer $\varepsilon$ for a test image. A classical technique for estimating a continuous parameter, such as $\varepsilon$, from training data is \emph{linear regression} (\cite{DudaR2001}). The method requires a pair $(\varepsilon_k,\ff_k)$ per training image $I_k$, where $\varepsilon_k$ is the ``optimal'' distortion for image $I_k$ and $\ff_k$ is a set of features extracted from $I_k$. Then the regression parameters $\ww$ can be estimated by solving the following objective function:
\begin{equation}
\label{eq:regression}
\ww^*=\argmin{\ww} \sum_{k} (\ww^T\ff_k-\varepsilon^*_k)^2.
\end{equation}
The distortion level $\varepsilon$ w.r.t. a new test image $I$ with its feature vector $\ff$ is given by $\varepsilon(\ff)\doteq {\ww^*}^T\ff$.

The features $\ff_k$ in \eqref{eq:regression} should be chosen to effectively model the statistics of the image, so that the relationship between $\varepsilon$ and $\ff_k$ is well approximated by the linear function $\varepsilon_k \approx \ww^T \ff_k$. A simple idea to define $\ff_k$ could consider how contrastive the regions in $I_k$ are. Intuitively, when the textures in $I_k$ are similar, such as in camouflage images, stronger sensitivity to contrast in patterns is required. Since computing the standard deviation of pixel intensities gives a measure of pattern contrast, 
we resize each $I_k$ with multiple scales, and define the features $\ff_k$ as the standard deviations of the pixel intensities at the multiple image resolutions.

Another issue in linear regression is that the classical model \eqref{eq:regression} is insufficient to accurately predict the distortion level for Algorithm \ref{alg:multi-resolution}. In particular, the discrepancy measure $d$ is only used to determine the optimal $\varepsilon^*$ for a training image. Segmentation results for other choices of $\varepsilon$ are not used in the regression. However, it is possible to better estimate the distortion $\varepsilon$ by taking into account the segmentation results around a neighborhood of the optimal distortion $\varepsilon^*$ in the training set.

For agglomerative image segmentation, the discrepancy measures that we use in this paper exhibit a simple behavior. Specifically, as $\varepsilon$ deviates from $\varepsilon^*$ in either direction, the discrepancy between the segmentation and the ground truth almost increases monotonically. This is because as $\varepsilon$ deviates from $\varepsilon^*$, it leads to over-segmentation or under-segmentation, both of which have larger discrepancies from the ground truth (see Figure \ref{fig:epsilon-vs-discrepancy}). Motivated by this observation, we approximate the discrepancy function $d$ by a convex quadratic form:
\begin{equation}
d(\mathcal{R}_\varepsilon(I_k),\mathcal{R}_g(I_k)) \approx a_k\varepsilon^2 + b_k\varepsilon + c_k, \quad \mbox{where } a_k>0.
\label{eq:quadratic-discrepancy}
\end{equation}
The parameters $(a_k, b_k, c_k)$ are then estimated by least squares fitting w.r.t. the pairs $(d_k, \varepsilon)$. The latter is attained by sampling the function $d(S_\varepsilon(I_k),S_g(I_k))$ at different $\varepsilon$'s.

Once we substitute \eqref{eq:quadratic-discrepancy} in \eqref{eq:optimal-epsilon-estimate} in combination with the linear model $\varepsilon = \ww^T\ff_k$, the objective function to recover the linear regression parameter $\ww^*$ is given by
\begin{equation}
\ww^* = \argmin{\ww}\sum_k a_k(\ww^Tf_k)^2 + b_k(\ww^Tf_k) + c_k.
\label{eq:regression3}
\end{equation} 
Since $a_k>0$ for all training images $I_k$, \eqref{eq:regression3} is an unconstrained convex program. Thus it has a closed-form solution:
\begin{equation}
\ww^* = -\frac{1}{2}(\sum_{k=1}^n a_k\ff_k\ff_k^T)^{-1}(\sum_k b_k\ff_k).
\end{equation}

Once $\ww^*$ is learned from the training data, the optimal distortion of the test image $I$ with its feature vector $\ff$ is predicted by $\varepsilon(\ff)={\ww^*}^T\ff$. We caution that, based on $\ww^*$, the prediction of the distortion parameter $\varepsilon(\ff_k)$ for each training image $I_k$ may not necessarily be the same as $\varepsilon_k^*$ selected from the ground truth $S_g(I_k)$. Nevertheless, the proposed solution ensures that the linear model minimizes the average discrepancy over the training data.

\section{Experiments}
\label{sec:experiments}

In this section, we conduct extensive evaluation to validate the performance of the TBES algorithm. The experiment is based on the publicly available Berkeley Segmentation Dataset (BSD) (\cite{MartinD2001-ICCV}). BSD is comprised of 300 natural images, which covers a variety of natural scene categories, such as portraits, animals, landscape, and beaches. The database is partitioned into a training set of 200 images and a testing set of 100 images. It also provides ground-truth segmentation results of all the images obtained by several human subjects. On average, five segmentation maps are available per image. Multiple ground truth allows us to investigate how human subjects agree with each other.

The implementation of the TBES algorithm and the benchmark scripts are available online at:

\url{http://perception.csl.illinois.edu/coding/image_segmentation/}

\subsection{Color Spaces and Compressibility}
The optimal coding length of textured regions of an image depends in part on the color space. We seek to determine the color space in which natural images are most compressible based on the proposed lossy compression scheme \eqref{eqn:total-coding-length}. It has been noted in the literature that the \emph{Lab} color space (also known as $L^*a^*b^*$) better approximates the perceptually uniform color metric \citep{JainA1989}. This has motivated some of the previous works \citep{YangA2008-CVIU,RaoS2009-ACCV}  to utilize such representation in methods for natural image segmentation. In order to check the validity of this assumption, particularly for our segmentation scheme by compressing texture, we perform a study on five color spaces that have been widely used in the literature, namely,  $Lab$, $YUV$, $RGB$, $XYZ$, and $HSV$.

We use the manually segmented training images in the Berkeley dataset to rank the compressibility of the 5 color spaces. Given a color space, for any image and corresponding segmentation, the number of bits required to encode texture information is computed by \eqref{eqn:old-coding}, with features constructed as in Section \ref{sec:Gaussianity}. The average coding length of an image is computed as the one over all ground-truth segmentation maps for that image. Finally, the average coding length of the dataset is computed over the entire images in the dataset.

We note that the volume of the pixel distribution (and thus the coding length) can change if the pixel values are rescaled. This means one color space can look more compressible by merely producing numbers in a smaller range, say $[0,1]$ as opposed to another which is in range $[0,255]$. In order to achieve a fair comparison, we normalize the feature vectors by scale factor $c$, which is constant across features from the same color space:
\begin{equation}
c=1/\sqrt{\bar{\lambda}_{\mbox{max}}}
\end{equation}
where $\bar{\lambda}_{\mbox{max}}$ is the average of the maximum eigenvalues of the feature covariance matrix over all regions and all images in the dataset.

\begin{figure}[ht!]
\centering
\includegraphics[width=3in]{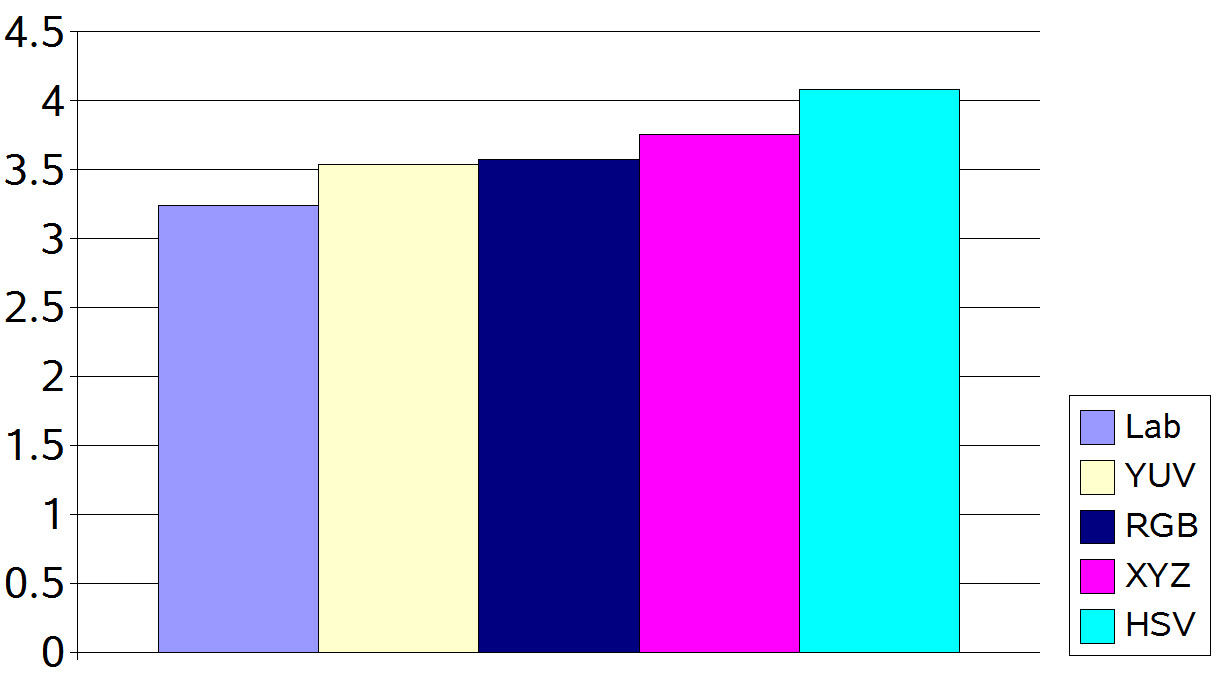}
\caption{\small Average coding length of an image in five representative color spaces.}
\label{fig:colorspace}   
\end{figure}

The average (normalized) coding lengths of five representative color spaces are shown in Figure \ref{fig:colorspace}. Among all the 5 color spaces examined,  $Lab$ has the shortest coding length. Therefore, in the rest of our experiments, input images are first converted to the  $Lab$ color space.

\subsection{Experimental Setup}
\label{sec:setup}

To quantitatively evaluate the performance of our method, we use  three metrics for comparing pairs of image segmentations: the {\em probabilistic Rand index} (PRI) (\cite{RandW1971-JASA}), the {\em variation of information} (VOI) (\cite{MeilaM2005-ICML}), and the {\em global F-measure} (\cite{ArbelaezP2006}):
\begin{enumerate} 
\item The {\em probabilistic Rand index} (PRI) is a classical metric that measures the probability that an arbitrary pair of samples have consistent labels in the two partitions. The PRI metric is in the range $[0,1]$, with higher values indicating greater similarity between two partitions. When used to adaptively choose $\varepsilon$ as described in Section \ref{sec:epsilon}, we use $\mbox{PRI}^C \doteq (1-\mbox{PRI})$.
\item The {\em variation of information} (VOI) measures the sum of information loss and information gain between the two clusterings, and thus it roughly measures the extent to which one clustering can explain the other. The VOI metric is nonnegative, with lower values indicating greater similarity.
\item The \emph{global F-measure} (GFM) is the harmonic mean of precision and recall, a pair of complimentary metrics for measuring the accuracy of the boundaries in an image segmentation given the ground truth boundaries. Precision measures the fraction of true boundary pixels in the test segmentation. Recall measures the fraction of ground-truth boundary pixels in the test segmentation. When used to adaptively choose $\varepsilon$, we use $\mbox{GFM}^C \doteq (1-\mbox{GFM})$.
\end{enumerate} 
In cases where we have multiple ground-truth segmentations, to compute the PRI or VOI measure for a test segmentation, we simply average the results of the metric between the test segmentation and each ground-truth segmentation. To compute the GFM measure from multiple ground-truth segmentations, we apply the same techniques used in \cite{ArbelaezP2009-CVPR}, which roughly aggregate the boundary precision and recall over all ground-truth images as an ensemble.
With multiple ground-truth segmentations for an image we can also estimate the human performance w.r.t. these metrics by treating each ground-truth segmentation as a test segmentation and computing the metrics w.r.t. the other ground-truth segmentations.

The adaptive $\varepsilon$ scheme in our method relies on the feature vector $\ff$ used in (\ref{eq:regression3}) as follows. The image $I$ is converted to grayscale and its size is rescaled by a set of specific factors. The standard deviation of pixel intensity of each rescaled image constitutes a component of the feature vector. Empirically, we have observed that using four scale factors , i.e., $\ff \in \mathbb{R}^4$, produces excellent segmentation results for our algorithm on the BSD database.

The parameters $(a_k, b_k, c_k)$ in the quadratic form in (\ref{eq:quadratic-discrepancy}) are estimated as follows. We sample $25 \leq \varepsilon \leq 400$ uniformly, in steps of 25 and compute the corresponding $d(S_\varepsilon(I_k),S_g(I_k))$ for each sample. This gives a set $\{(d_{k,n}, \varepsilon_{k,n})\}_{n=1}^{16}$ for an image $I_k$. We use this set to estimate $(a_k, b_k, c_k)$ by least squares method. 

\subsection{Results}
\label{sec:results}

We quantitatively compare the performance of our method with six \emph{publicly available} image segmentation methods, namely, \emph{Mean-Shift} (MS) by \cite{ComaniciuD2002-PAMI}, \emph{Markov Chain Monte Carlo} (MCMC) by \cite{TuZ2002-PAMI}, \emph{F\&H} by \cite{FelzenszwalbP2004-IJCV}, \emph{Multiscale NCut} (MNC) by \cite{CourT2005-CVPR}, {\em Compression-based Texture Merging} (CTM) by \cite{YangA2008-CVIU}, and \emph{Ultrametric Contour Maps} (UCM) by \cite{ArbelaezP2009-CVPR}, respectively. We refer to our method as ``TBES''. The user-defined parameters of these methods have been tuned by the training subset of each dataset to achieve the best performance w.r.t. each segmentation index. Then, the performance of each method is evaluated based on the test subset.

\begin{figure}[htbp]
\centering
\subfigure[Animals]{
\begin{minipage}{.18 \textwidth}
\centering
\includegraphics[width=.99\textwidth]{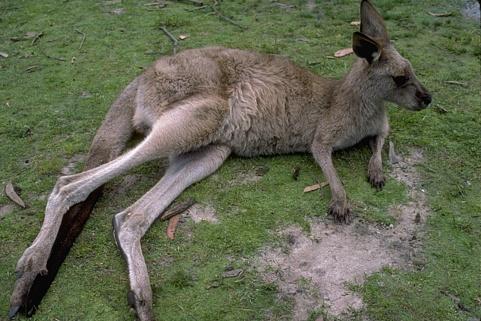} 
\includegraphics[width=.99\textwidth]{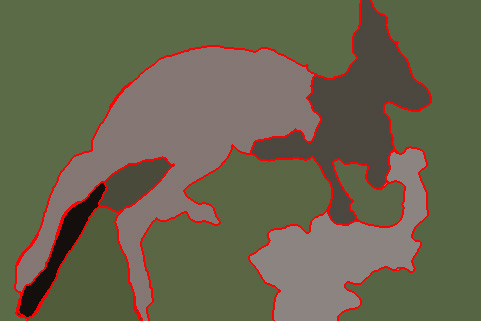}
\includegraphics[width=.99\textwidth,clip=true,viewport=1 264 321 481]{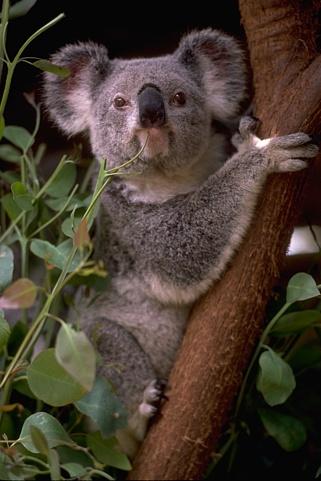} \\
\includegraphics[width=.99\textwidth,clip=true,viewport=1 264 321 481]{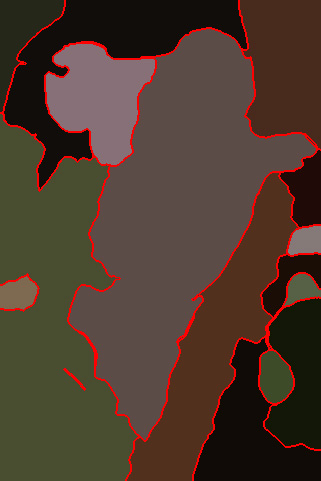}
\includegraphics[width=.99\textwidth]{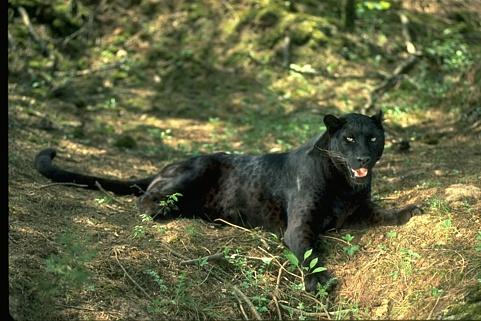} 
\includegraphics[width=.99\textwidth]{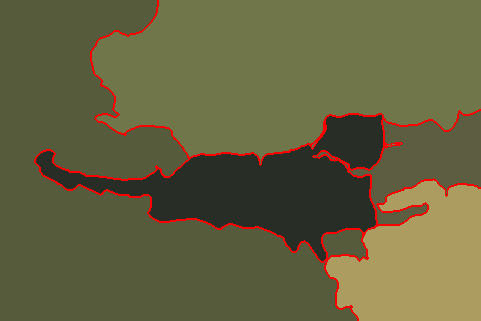}
\end{minipage}}
\subfigure[Buildings]{
\begin{minipage}{.18 \textwidth}
\includegraphics[width=.99\textwidth]{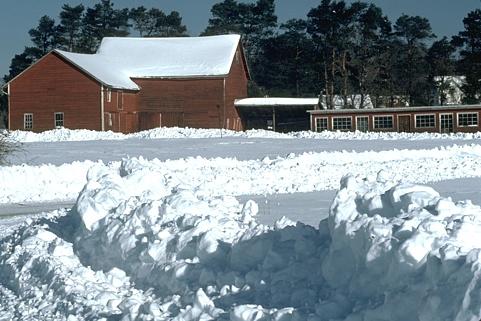} 
\includegraphics[width=.99\textwidth]{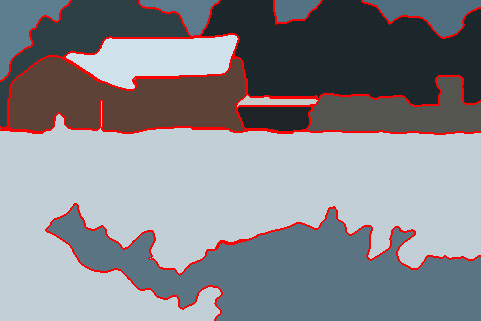}
\includegraphics[width=.99\textwidth,clip=true,viewport=1 214 321 431]{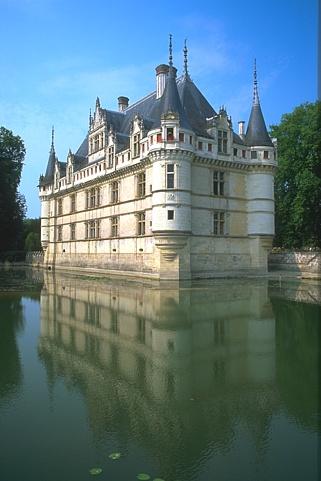} 
\includegraphics[width=.99\textwidth,clip=true,viewport=1 214 321 431]{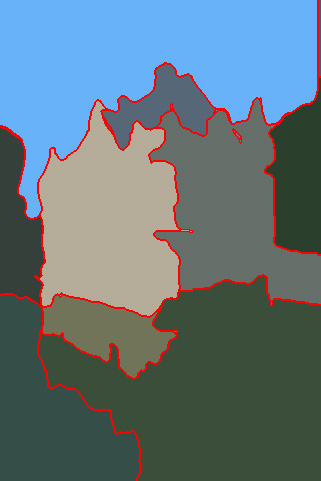}
\includegraphics[width=.99\textwidth,clip=true,viewport=1 264 321 481]{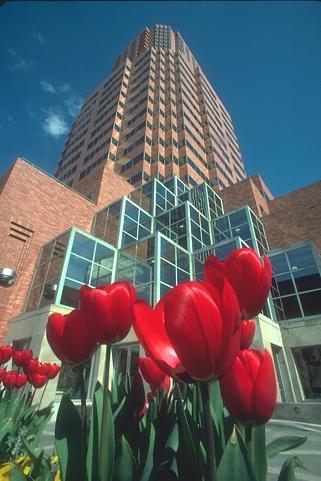} 
\includegraphics[width=.99\textwidth,clip=true,viewport=1 264 321 481]{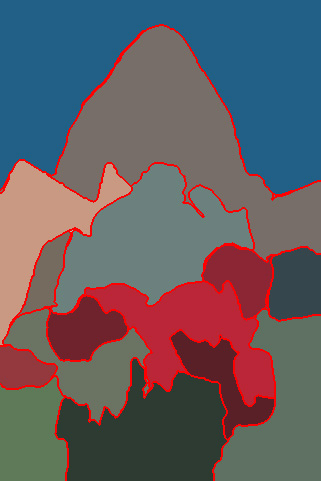}
\end{minipage}}
\subfigure[Landscape]{
\begin{minipage}{.18 \textwidth}
\includegraphics[width=.99\textwidth]{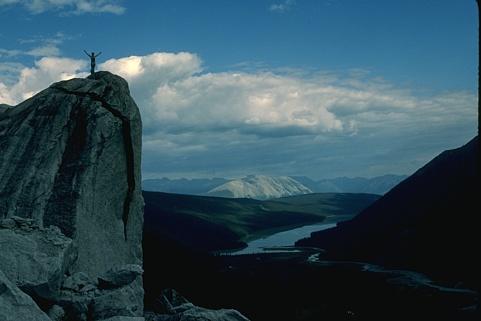} 
\includegraphics[width=.99\textwidth]{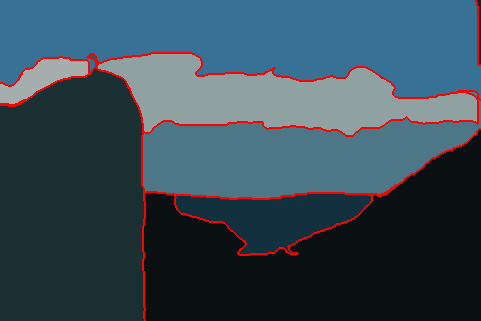}
\includegraphics[width=.99\textwidth]{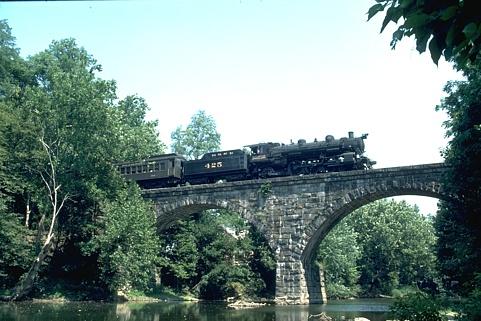} 
\includegraphics[width=.99\textwidth]{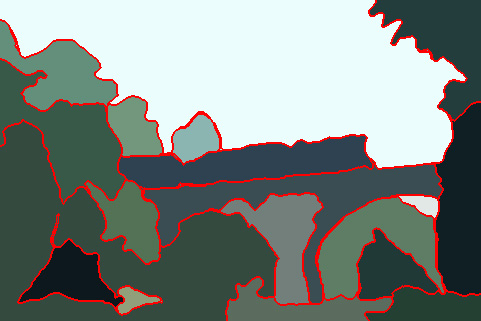}
\includegraphics[width=.99\textwidth]{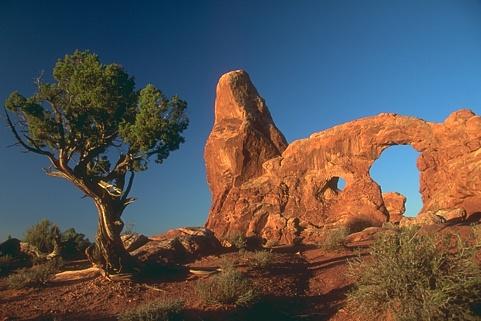} 
\includegraphics[width=.99\textwidth]{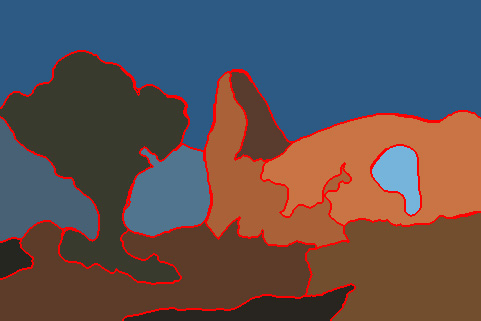}
\end{minipage}}
\subfigure[People]{
\begin{minipage}{.18 \textwidth}
\includegraphics[width=.99\textwidth]{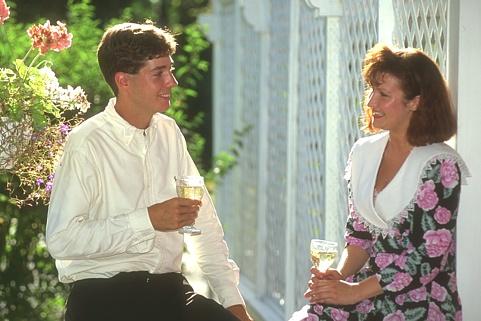} 
\includegraphics[width=.99\textwidth]{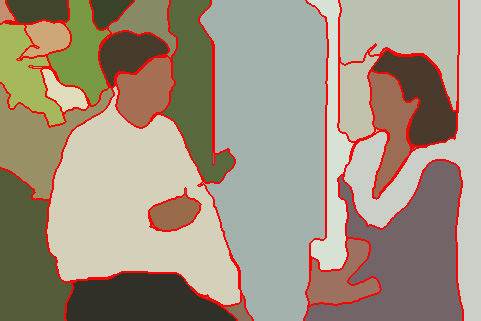}
\includegraphics[width=.99\textwidth,clip=true,viewport=1 194 321 411]{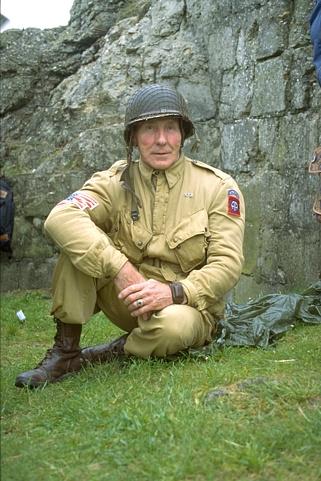} 
\includegraphics[width=.99\textwidth,clip=true,viewport=1 194 321 411]{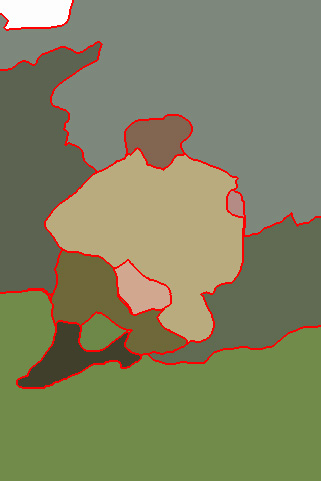}
\includegraphics[width=.99\textwidth]{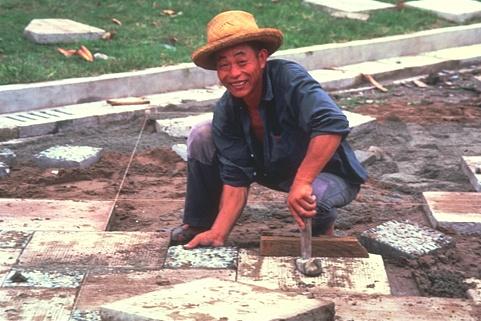} 
\includegraphics[width=.99\textwidth]{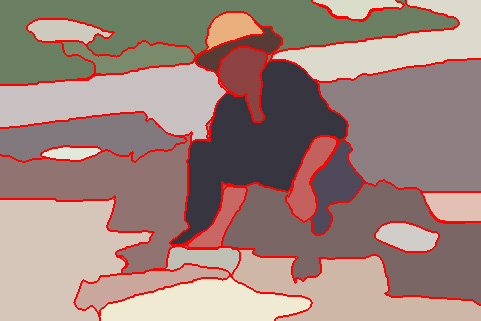}
\end{minipage}}
\subfigure[Water]{
\begin{minipage}{.18 \textwidth}
\includegraphics[width=.99\textwidth]{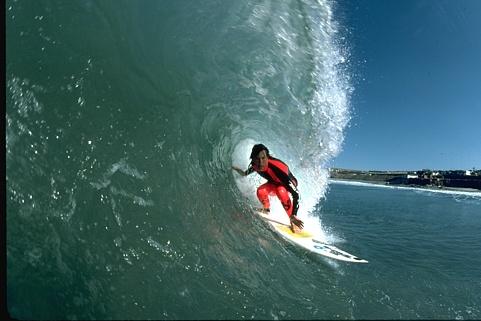} 
\includegraphics[width=.99\textwidth]{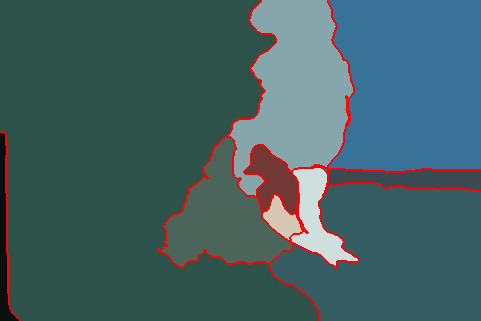}
\includegraphics[width=.99\textwidth]{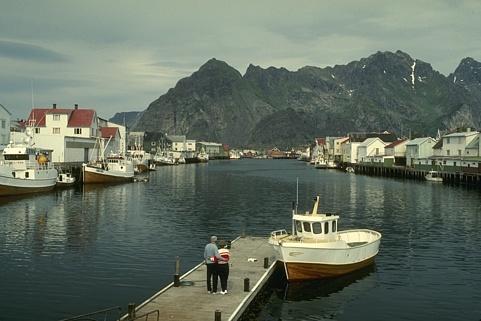} 
\includegraphics[width=.99\textwidth]{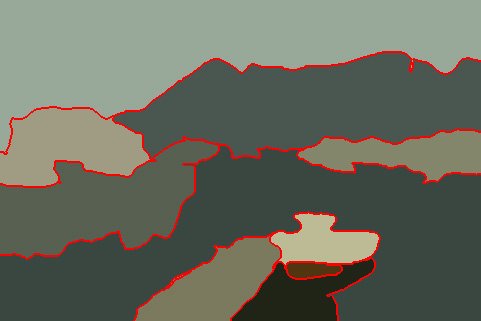}
\includegraphics[width=.99\textwidth]{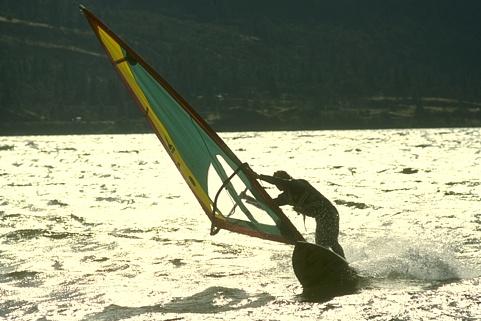} 
\includegraphics[width=.99\textwidth]{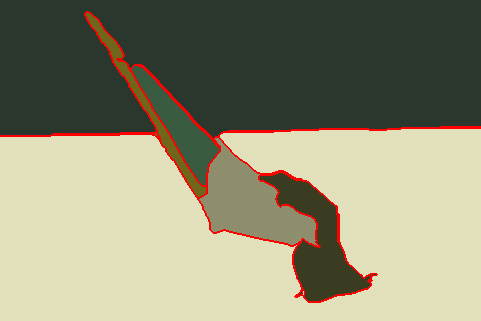}
\end{minipage}}
\caption{\small Representative segmentation results (in color) of the TBES algorithm on various image categories from BSD.  For each image pair, the top is the original input image, and the bottom is the segmentation result where each texture region is rendered by its mean color.  The distortion $\varepsilon$ was chosen adaptively to optimize PRI.}
\label{fig:berkeley-images}   
\end{figure}

Table \ref{tab:BSD-quant} shows the segmentation accuracy of the TBES algorithm compared to the human ground truth and the other six algorithms.\footnote{The quantitative performance of several existing algorithms was also evaluated in a recent work (\cite{ArbelaezP2009-CVPR}), which was proposed roughly at the same time as this paper. The reported results therein generally agree with our findings.} In addition to the computational methods, multiple ground truth segmentations in BSD allow us to estimate the human performance w.r.t. these metrics. This was achieved by treating each ground-truth segmentation as a test segmentation and computing the metrics w.r.t. the other ground-truth segmentations. To qualitatively inspect the segmentation, Figure \ref{fig:berkeley-images} illustrates some representative results.
\begin{table}[htbp]
\caption{\small Comparison on the BSD using PRI, VOI, and GFM indices. For PRI and GFM, higher values indicate better segmentation; for VOI, lower values indicate better segmentation.}
\centering
\begin{small}
\begin{tabular}{|c||c||c|c|c|c|c|c|c|}
\hline
BSD & Human & TBES  & MS & MCMC & F\&H & MNC & CTM & UCM \\
\hline
\hline
PRI &
0.868 & 
\bf{0.807} & 
0.772 & 
0.768 & 
0.770 & 
0.742 & 
0.755 & 
0.796 
\\
\hline 
VOI &
1.163 & 
\bf{1.705} & 
2.004 & 
2.261 & 
2.188 & 
2.651 & 
1.897 & 
1.715 
\\
\hline 
GFM &
0.787 & 
0.647 & 
0.600 & 
0.467 & 
0.579 & 
0.590 & 
0.595 & 
\bf{0.706} 
\\
\hline
\end{tabular}
\end{small} 
\label{tab:BSD-quant}
\end{table}

Among all the algorithms in Table \ref{tab:BSD-quant}, TBES achieves the best evaluations w.r.t. PRI and VOI. The second best algorithm reported is the UCM algorithm. It is also worth noting that there seems to be a large gap in terms of VOI between all the algorithm indices and the human index (e.g., 1.705 for TBES versus 1.163 for human). With respect to GFM, UCM achieves the best performance, which is mainly due to the fact that UCM was designed to construct texture regions from the hierarchies of (strong) image contours and edges. In this category, our algorithm still achieves the second best performance, largely exceeding the indices posted by the rest of the algorithms in the literature.

\begin{figure}[ht!]
\centering
\subfigure[Camouflage]{
\begin{minipage}{.2 \textwidth}
\includegraphics[width=.99\textwidth]{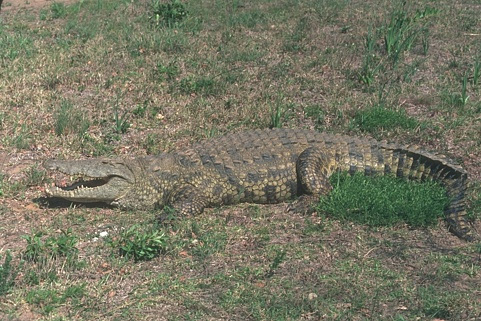}
\includegraphics[width=.99\textwidth]{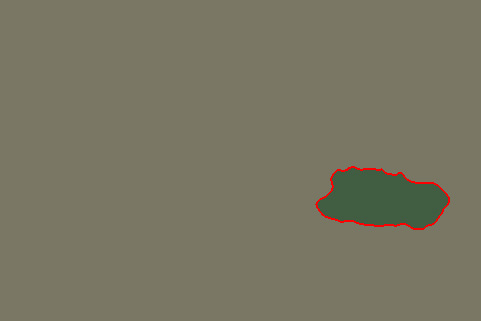}
\includegraphics[width=.99\textwidth]{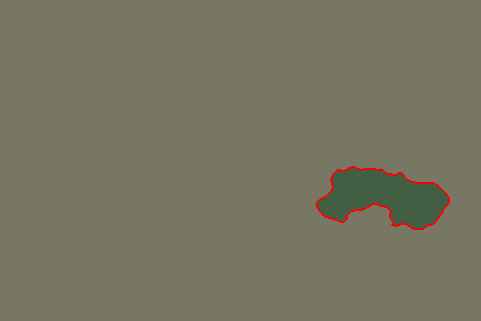}
\end{minipage}}
\subfigure[Shadows]{
\begin{minipage}{.2 \textwidth}
\centering
\includegraphics[width=.99\textwidth]{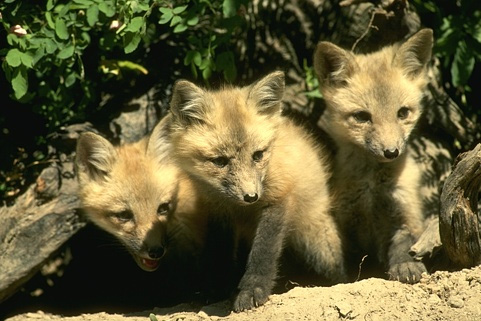}
\includegraphics[width=.99\textwidth]{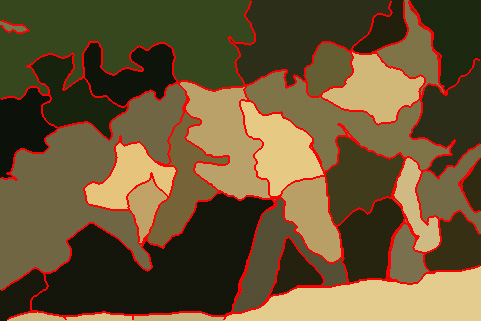}
\includegraphics[width=.99\textwidth]{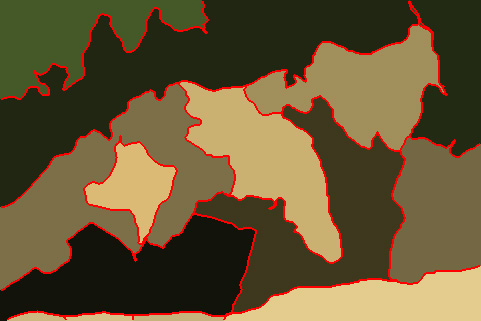}
\end{minipage}}
\subfigure[Non-Gaussian]{
\begin{minipage}{.2 \textwidth}
\includegraphics[width=.99\textwidth]{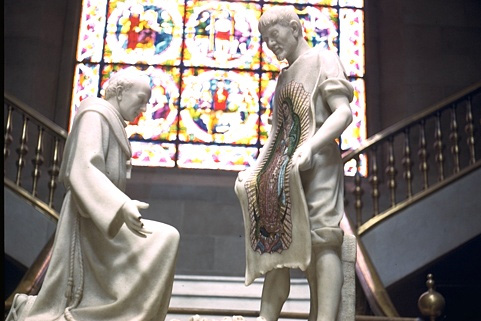}
\includegraphics[width=.99\textwidth]{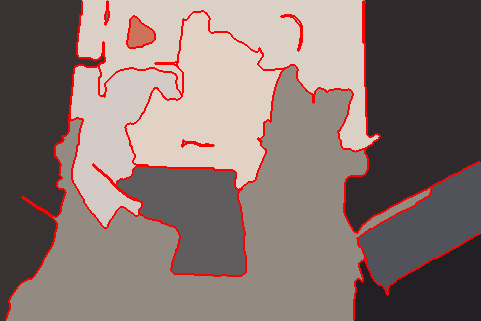}
\includegraphics[width=.99\textwidth]{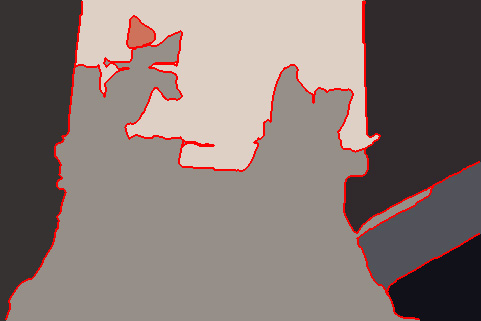}
\end{minipage}}
\subfigure[Thin Regions]{
\begin{minipage}{.2 \textwidth}
\includegraphics[width=.99\textwidth]{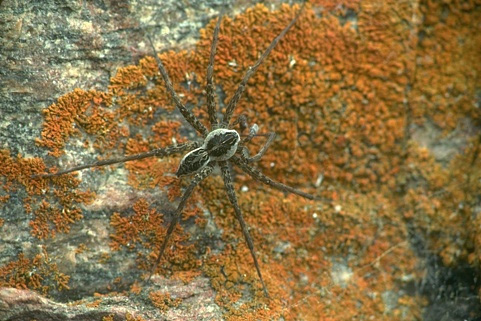}
\includegraphics[width=.99\textwidth]{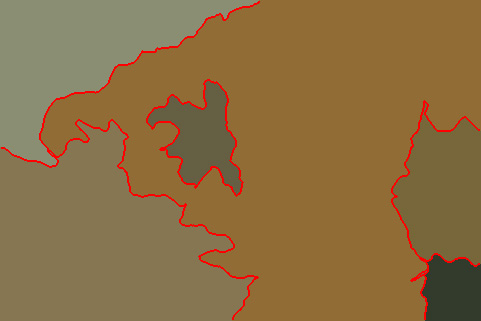}
\includegraphics[width=.99\textwidth]{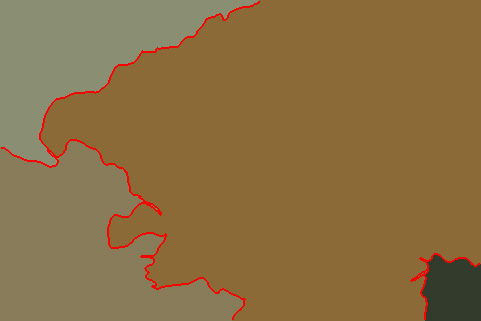}
\end{minipage}}
\caption{\small Examples from BSD (in color) where TBES algorithm failed obtaining a reasonable segmentation. {\bf Top:} Original input images. {\bf Middle:} Segmentation w.r.t PRI. {\bf Bottom:} Segmentation w.r.t VOI.}
\label{fig:BSD-Failure}
\end{figure}

Finally, we briefly discuss a few images on which our method fails to achieve a reasonable segmentation. The examples are shown in Figure \ref{fig:BSD-Failure}. The main causes for visually inferior segmentation are camouflage, shadows, non-Gaussian textures, and thin regions:
\begin{enumerate}
\item It is easy to see that the texture of animal camouflages is deliberately chosen to be similar to the background texture. The algorithm falls behind humans in this situation, arguably, because human vision can \emph{recognize} the holistic shape and texture of the animals based on experiences.
\item As shades of the same texture may appear very different in images, TBES may break up the regions into more or less the same level of shade.
\item Some patterns in natural images do not follow the Gaussian texture assumption. Examples include geometric patterns such as lines or curves.
\item Thin regions, such as spider's legs, are problematic for TBES for two reasons. First, it has trouble to properly form low-level superpixels used as the initialization of our method. Second, large enough windows which can better capture the statistics of the texture can barely fit into such thin regions. Consequently, texture estimation at these regions is ill-conditioned and unstable.
\end{enumerate}

To realize whether these problems are unique to our method or are more universal, we have investigated similar problematic cases with the other methods reported here (\cite{ComaniciuD2002-PAMI,TuZ2002-PAMI,FelzenszwalbP2004-IJCV,CourT2005-CVPR,YangA2008-CVIU,ArbelaezP2009-CVPR}). None of the methods were able to handle camouflage very well. Shadows are challenging for these methods as well. However, we observe that UCM performs relatively better in this case. For geometric patterns, CTM seems to be slightly better than others, but still is an over-segmentation. In the category of thin regions, all algorithms performed very poorly, but mean-shift is better by, for example, roughly picking up some of the spider's legs. It is further worth pointing out an interesting observation about PRI versus VOI that the former prefers over-segmentation and the latter prefers under-segmentation (as shown in Figure \ref{fig:BSD-Failure}).
 
\section{Conclusion}
\label{sec:conclusion}

We have proposed a novel method for natural image segmentation. The algorithm uses a principled information-theoretic approach to combine cues of image texture and boundaries. In particular, the texture and boundary information of each texture region is encoded using a Gaussian distribution and adaptive chain code, respectively. The partitioning of an image is achieved by an agglomerative clustering process applied to a hierarchy of decreasing window sizes. Based on the MDL principle, the optimal segmentation of the image is defined as the one that minimizes its total coding length. As the lossy coding length function also depends on a distortion parameter that determines the coarseness of the segmentation, we have further proposed an efficient linear regression method to learn the optimal distortion parameter from a set of training images when provided by the user. Our experiments have validated that the new algorithm outperforms other existing methods in terms of region-based segmentation indices, and is among the top solutions in terms of contour-based segmentation indices. To aid peer evaluation, the implementation of the algorithm and the benchmark scripts have been made available on our website.

\bibliographystyle{IEEE}

\end{document}